\documentclass{article}

    \PassOptionsToPackage{numbers, compress}{natbib}

\usepackage[final]{neurips_2021}

\usepackage[utf8]{inputenc} 
\usepackage[T1]{fontenc}    
\usepackage{hyperref}       
\usepackage{url}            
\usepackage{booktabs}       
\usepackage{amsfonts}       
\usepackage{nicefrac}       
\usepackage{microtype}      
\usepackage{xcolor}         
\usepackage{comment}

\usepackage{graphicx}       
\usepackage{caption}
\usepackage{algorithm}
\usepackage{algorithmicx}
\usepackage{algpseudocode}
\usepackage{amsmath}
\usepackage{amsthm}
\usepackage{wrapfig}
\usepackage{amssymb}
\usepackage{multirow}
\usepackage{multicol}
\usepackage{enumitem}
\usepackage{caption}
\usepackage{subfigure}

\newcommand{\etal}{\textit{et al}.~}

\newcommand{\ieno}{\textit{i}.\textit{e}.}

\newcommand{\egno}{\textit{e}.\textit{g}.} 

\newcommand{\tcb}{}
\newcommand{\tcr}{}

\newcommand{\ourname}{PlayVirtual}
\newcommand{\bmark}{$^{\dagger}$}

\title{PlayVirtual: Augmenting Cycle-Consistent Virtual Trajectories for Reinforcement Learning}

\author{%
  Tao Yu$^{1}$\thanks{This work was done when Tao Yu was an intern at Microsoft Research Asia.} ~
  Cuiling Lan$^{2}$\thanks{Corresponding Author.} ~
  Wenjun Zeng$^{2}$ ~
  Mingxiao Feng$^{1}$ ~
  Zhizheng Zhang$^{2}$ ~
  Zhibo Chen$^{1}$\bmark \\ $^{1}$University of Science and Technology of China~~~~$^{2}$Microsoft Research Asia \\ 
  \texttt{yutao666@mail.ustc.edu.cn, \{culan,wezeng\}@microsoft.com} \\ \texttt{fmxustc@mail.ustc.edu.cn, zhizzhang@microsoft.com, chenzhibo@ustc.edu.cn}
  }

\begin{document}

\maketitle

\begin{abstract}
Learning good feature representations is important for deep reinforcement learning (RL).
However, with limited experience, RL often suffers from data inefficiency for training. 
For un-experienced or less-experienced trajectories (\ieno, state-action sequences), the lack of data limits the use of them for better feature learning.
In this work, we propose a novel method, dubbed \ourname, which augments cycle-consistent virtual trajectories to enhance the data efficiency for RL feature representation learning.
Specifically, \ourname~predicts future states \tcr{in the latent space} based on the current state and action by a dynamics model and then predicts the previous states by a backward dynamics model, which forms a trajectory cycle. 
Based on this, we augment the actions to generate a large amount of virtual state-action trajectories. 
Being free of groudtruth state supervision, we enforce a trajectory to meet the cycle consistency constraint, which can significantly enhance the data efficiency. 
We validate the effectiveness of our designs on the Atari and DeepMind Control Suite benchmarks. \tcr{Our method achieves the state-of-the-art performance on both benchmarks.}
   
\end{abstract}

\section{Introduction}
Deep reinforcement learning (RL) combines the powerful representation capacity of deep neural networks and the notable advantages of RL for solving sequential decision-making problems. It has made great progress in many complex control tasks such as video games \cite{mnih2015human,vinyals2019grandmaster,berner2019dota}, and robotic control \cite{kalashnikov2018qt,zhang2019solar,nagabandi2018learning}.
Despite the success of deep RL, it faces the challenge of data/sample inefficiency when learning from high-dimensional observations such as image pixels 
from limited experience \cite{lake2017building,Kaiser2020Model,laskin2020curl}. 
Fitting a high-capability feature encoder using only scarce reward signals is data inefficient and prone to suboptimal convergence~\cite{yarats2019improving}. Humans can learn to play Atari games in several minutes, while RL agents need millions of interactions \cite{Tsividis2017HumanLI}.
However, collecting experience in the real world is often expensive and time-consuming. One may need several months to collect interaction data for robotic arms training \cite{kalashnikov2018qt} or be troubled by collecting sufficient patient data to train a healthcare agent \cite{yu2019reinforcement}.
Therefore, from another perspective, making efficient use of limited experience for improving data efficiency becomes vital for RL.

Many methods improve data efficiency by introducing auxiliary tasks with useful self-supervision to learn compact and informative feature representations, which better serves policy learning. 
Previous works have demonstrated that good auxiliary supervision can significantly improve agent learning, like leveraging image reconstruction \cite{yarats2019improving}, the prediction of future states \cite{Shelhamer2017LossII,guo2020bootstrap,lee2020slac,schwarzer2021dataefficient}, maximizing Predictive Information 
\cite{oord2018representation,anand2019stdim,mazoure2020deep,stooke2020decoupling,lee2020predictive}, or promoting discrimination through contrastive learning
\cite{laskin2020curl,zhu2020masked,liu2021returnbased,Kipf2020Contrastive}. Although the above methods have been proposed to improve the data efficiency of RL, 
the limited experience still hinders the achievement of high performance. 
For instance, the current state-of-the-art method SPR~\cite{schwarzer2021dataefficient} only achieves about $40\%$ of human level on Atari~\cite{bellemare2013arcade} when \tcb{using data from 100k interactions with the environment}. 
Some methods improve data efficiency by applying modest image augmentation (\ieno, transformations of the input images like 
random shifts and intensity)
\cite{laskin2020reinforcement,yarats2021image}.
Such perturbation on images improves the diversity of appearances of the input images. However, it cannot enrich the experienced trajectories (state-action sequences) in training and thus 
\textbf{the deep networks are still deficient in experiencing/ingesting flexible/diverse trajectories}.   

\begin{figure*}[t]
    \vspace{0pt}
    \begin{center}
        \subfigure[]{\includegraphics[width=0.49\hsize]{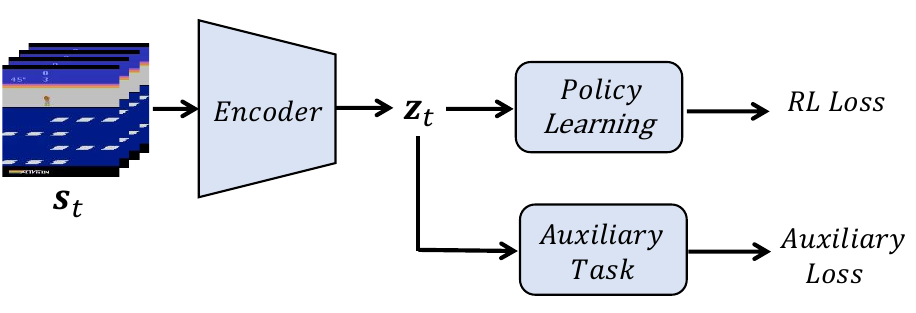}\label{fig:framework-a}}  
        \subfigure[]{\includegraphics[width=0.5\hsize]{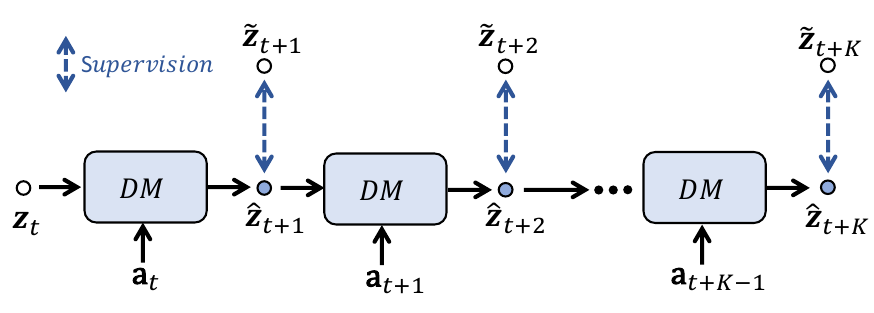}\label{fig:framework-b}} \\ \vspace{-5pt}
        \subfigure[]{\includegraphics[width=0.68\hsize]{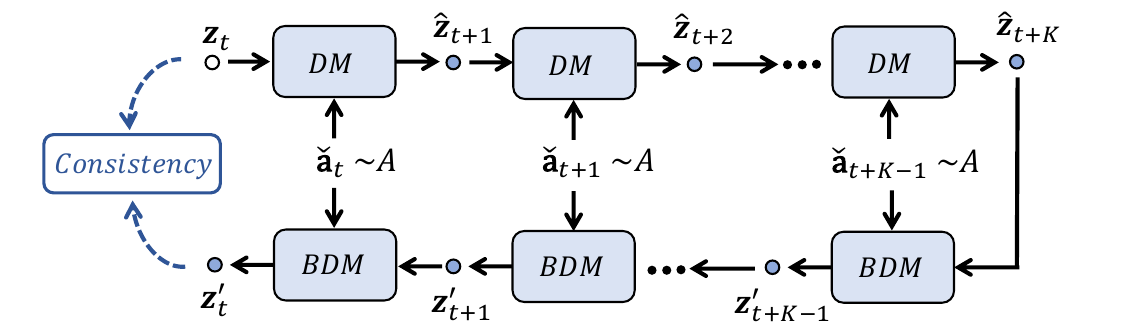}\label{fig:framework-c}} \vspace{-5pt}
    \end{center}
    \caption{Illustration of the main pipeline of our method. (a) A glance at the overall framework which consists of an encoder for learning the latent state representation $\mathbf{z}_t$, a policy learning head,
    and our auxiliary task module. The auxiliary task module consumes a real trajectory as shown in (b) and an augmented virtual trajectory as shown in (c), respectively. In (b), we train the dynamics model (DM) to be predictive of the future state based on the input state and action, with the supervision from the future state. 
    To enhance data efficiency, as shown in (c), we augment the actions to generate a virtual trajectory formed by a forward and a backward trajectory. 
    Particularly, the forward state-action trajectory is obtained based on the current state $\mathbf{z}_t$, the DM and \tcr{a sequence of augmented/generated actions} (\egno, for $K$ steps).
    Similarly, based on the predicted future state $\hat{\mathbf{z}}_{t+K}$, a backward dynamics model (BDM), and that sequence of augmented actions, we obtain the backward state-action trajectory. \emph{For the virtual trajectory, we add the consistency constraint on the current state} $\mathbf{z}_t$ \emph{and the predicted current state} $\mathbf{z}'$ \emph{for optimizing the feature representations}.} 
    \label{fig:framework}
\end{figure*}

In this work, to address the above problem, we propose a new method dubbed \ourname, which augments cycle-consistent virtual trajectories to improve data efficiency. 
Particularly, we predict future states \tcr{in the latent space} by a dynamics model in the forward prediction (trained using real trajectories to predict the future state based on the current state and action) and then predict the previous \tcr{states} by a backward dynamics model, forming a loop.
In this way, we can augment the actions to generate a large amount of virtual/fictitious state-action transitions for feature representation training, with the self-supervised cycle consistency constraint (which is a necessary/desirable condition for a good feature encoder).
Note that such design is free of groundtruth supervision. Our augmentation to generate abundant virtual trajectories can significantly enhance the data efficiency. 
As illustrated in Figure \ref{fig:framework}, on top of a baseline RL framework, we introduce our \ourname~for augmenting state-action virtual trajectories with cycle consistency constraint. The dynamics model infers the future states recurrently based on the current state and a set of randomly sampled actions, and the backward dynamics model predicts the previous states according to the predicted future state and those sampled/augmented actions. We enforce the backwardly predicted state of the current time step to be similar to the original current state to meet the cycle consistency constraint. 

We summarize the contributions of this work as follows:
\begin{itemize}[leftmargin=*,noitemsep,nolistsep]
    \item We pinpoint that augmenting the experience of RL in terms of trajectories is important for feature representation learning in RL (which is a sequential decision-making problem) to enhance data efficiency. 
    To our best knowledge, we are the first to generate virtual trajectories (experience) for boosting feature learning. 
	\item We propose a practical method \ourname~for augmenting virtual trajectories under self-supervised cycle consistency constraint, which significantly improves the data efficiency.
\end{itemize}

We demonstrate the effectiveness of our \ourname~on  discrete control benchmark Atari~\cite{bellemare2013arcade}, and continuous control benchmark DMControl Suite \cite{tassa2018deepmind}, where our \ourname~achieves the best performance on both benchmarks.

\section{Related Work}

\subsection{Data/Sample-efficient RL and Representation Learning for RL}
Learning from visual observations is a fundamental yet challenging problem in RL.
In practice, collecting experience is expensive and time-consuming. Thus, in general only limited experience is available for training RL agent, which results in the difficulty in fitting a high-capability feature encoder, \ieno, learning powerful feature representations.
Therefore, data-efficient RL has attracted a lot of attention and many methods are designed to make efficient use of limited experience for improving data efficiency.
These methods can be grouped into three categories. (i) Auxiliary task based methods introduce auxiliary task to help representation learning of the states \cite{
jaderberg2016reinforcement, yarats2019improving, Shelhamer2017LossII, guo2020bootstrap, lee2020slac, schwarzer2021dataefficient, oord2018representation, anand2019stdim, mazoure2020deep, stooke2020decoupling, lee2020predictive, laskin2020curl, zhu2020masked, liu2021returnbased}. 
(ii) Data augmentation based methods increase the diversity of image appearance through data augmentation~\cite{yarats2021image,laskin2020reinforcement,schwarzer2021dataefficient}. But they do not augment virtual actions.
(iii) World model based methods explicitly model the environment in order to enable the planning or promote the policy learning \cite{hafner2019learning,Hafner2020Dream,Kaiser2020Model}. 
We only focus on the first two categories since the third one is not specifically designed to enhance the efficiency of feature representation learning.

In recent years, unsupervised representation learning has made significant progress in natural language processing \cite{devlin2018bert,liu2019roberta} and computer vision \cite{oord2018representation,he2020momentum,chen2020simple,caron2020unsupervised,grill2020byol}.
It aims to learn generic and discriminative feature representations without groudtruth supervision, \ieno, by introducing some unsupervised auxiliary tasks. 
In RL, a good state representation removes the redundancy and noise elements from the original high-dimensional state, and reshapes the original state space into a compact feature space. 
Recently, many works explore representation learning in RL and have shown promising performance improvement.
UNREAL \cite{jaderberg2016reinforcement} introduces a number of auxiliary tasks such as reward prediction. 
Yarats \etal \cite{yarats2019improving} introduce an auxiliary reconstruction loss to aid feature representation learning. 
Considering the ability to model what will happen next is necessary for success on many RL tasks \cite{ha2018world,lee2020predictive}, some works train agents to be predictive of the future states.
\cite{Shelhamer2017LossII}, PBL \cite{guo2020bootstrap}, SLAC \cite{lee2020slac} and SPR \cite{schwarzer2021dataefficient} explicitly predict the future states by modeling dynamics.
Similarly, CPC \cite{oord2018representation}, ST-DIM \cite{anand2019stdim}, DRIML \cite{mazoure2020deep}, ATC \cite{stooke2020decoupling} and PI-SAC \cite{lee2020predictive} maximize the mutual information between the current state and the future state by using InfoNCE \cite{oord2018representation,stooke2020decoupling}, Deep InfoMax \cite{hjelm2018learning,anand2019stdim,mazoure2020deep}, or Conditional Entropy Bottleneck \cite{fischer2020conditional,lee2020predictive}.
Some works exploit contrastive learning to learn discriminative representations~\cite{laskin2020curl,zhu2020masked,liu2021returnbased,Kipf2020Contrastive}. CURL \cite{laskin2020curl} extracts high-level features from raw pixels using contrastive learning, by encouraging the similarity between the augmentations of the same image and the dissimilarity between different images.

Inspired by the success of data augmentation in computer vision, DrQ \cite{yarats2021image} and RAD \cite{laskin2020reinforcement} explore the effectiveness of data augmentation in RL and show that increasing the diversity of the training images by simple image augmentation (such as random cropping) can improve the data-efficiency. In SPR \cite{schwarzer2021dataefficient}, besides the prediction of its own latent state representations multiple steps into the future, they improve performance by adding data augmentation on the future input image as the future state supervision.
However, all these methods train the encoder using the real interaction transitions. There is a lack of an efficient mechanism to generate reliable state-action pair transition trajectories for better training the feature encoder.

In this paper, we propose a method, dubbed \ourname, which enables the augmentation of trajectories with unsupervised cycle consistency constraint for training the feature representation. 
This effectively enhances the data efficiency and our trajectory augmentation is conceptually orthogonal/complementary to the previous augmentation or auxiliary task based methods.

\subsection{Cycle Consistency}

Many works have explored the high-level idea of cycle consistency to address different challenges in various tasks such as image-to-image translation \cite{zhu2017unpaired,yi2017dualgan,kim2017learning}, image matching \cite{zhou2015flowweb,zhou2015multi,zhou2016learning}, feature representation learning \cite{wang2019learning,dwibedi2019temporal,kong2020ccl}.
For image-to-image translation, CycleGAN \cite{zhu2017unpaired} introduces the cycle consistency constraint to Generative Adversarial Networks \cite{goodfellow2014generative} to remove the requirement of groudtruth paired images for training, by enforcing the back-translated image to be the same as the original one.
Zhang \etal \cite{zhang2020learning} learn the correspondences that align dynamic robot behavior across two domains using cycle consistency: 
given observations in time $t$, the future prediction in time $t+1$ should be consistent across two domains under the consistent action taken. 
The purpose is to enable a direct transfer of the policy trained in one domain to the other without any additional fine-tuning.
To learn visual correspondence from unlabeled video, Wang \etal \cite{wang2019learning} propose to track a target backward and then forward to meet a temporal cycle consistency on the feature representations, using the inconsistency between the start and end points as the loss function.
Kong \etal \cite{kong2020ccl} propose cycle-contrastive learning for learning video representations, which is designed to find correspondences across frames and videos considering the contrastive representation in their domains respectively, where the two domain representations form a video-frame-video cycle.

Different from the above works, in order to improve data efficiency in RL, we propose to augment virtual state-action trajectories to enrich the "experience" of the feature encoder for representation learning. To ensure the reasonbleness/correctness of the generated transitions/trajectories and make use of them, we take the necessary condition of a good trajectory, \ieno, satisfying the cycle consistency of the trajectory, as a constraint to optimize the network. 

\section{\tcb{Cycle-Consistent Virtual Trajectories for Representation Learning in RL}} \label{sec:approach}

\subsection{Background}
We consider reinforcement learning (RL) in the standard Markov Decision Process (MDP) setting where an agent interacts with the environment in episodes. We denote the state, the action taken by the agent and the reward received at timestep $t$ in an episode as $\mathbf{s}_t$, $\mathbf{a}_{t}$, and $r_t$, respectively. We aim to train an RL agent whose expected cumulative reward in each episode is maximized. 

With the observation being high-dimensional short video clip at each timestep, the powerful representation capability of deep neural networks for encoding state and a strong RL algorithm contribute to the success of an RL agent. Similar to \cite{laskin2020curl}, we use the widely adopted RL algorithm Rainbow \cite{hessel2018rainbow} for discrete control benchmarks (\egno, Atari \cite{bellemare2013arcade}) and Soft Actor Critic (SAC) \cite{haarnoja2018soft} for continuous control benchmarks (\egno, DMControl Suite \cite{tassa2018deepmind}). 
Following SPR \cite{schwarzer2021dataefficient}, we introduce a dynamics model (DM) to predict the future latent states multiple steps, which enables a forward state-action trajectory. We take SPR~\cite{schwarzer2021dataefficient} as our baseline scheme.

\vspace{-0pt}
\subsection{Overall Framework}

Considering the data efficiency in RL with limited experience, we propose a method named \ourname~to efficiently improve the feature representation learning of RL. Our key idea is to augment the actions to generate \emph{virtual} state-action trajectories for boosting the representation learning of the encoder. Particularly, we eliminate the need of groudtruth trajectory supervision for the augmented sequences by using a cycle consistency constraint, which thus enhances data efficiency in training.

Figure~\ref{fig:framework} illustrates the main pipeline of our framework (with some details not presented for clarity). As shown in (a), it consists of an encoder which encodes the input observation $\mathbf{s}_t$ into low-dimensional latent state representation $\mathbf{z}_t$, an RL policy learning head (Rainbow~\cite{hessel2018rainbow} or SAC~\cite{haarnoja2018soft}), and our auxiliary task module. Particularly, as shown in (c), our auxiliary task module consists of a dynamics model (DM) which predicts future latent state based on the current state and the action, and a backward dynamics model (BDM) for backward state prediction. Following SPR~\cite{schwarzer2021dataefficient}, the DM is trained with the real state-action trajectory under the supervision of the future state (see (b)) to assure its capability of generating "correct" state transition. However, under limited experience, the encoder has few opportunities to be trained by those un-experienced or less-experienced state-action trajectories, which should be important to enhance data efficiency. To address this problem, as illustrated in (c), we add a BDM which predicts the previous state based on the current state and the previous action. Together with the DM,  
the forward predictions and backward predictions form a cycle/loop, where the current state and the backwardly predicted current state are expected to be the same. 
Particularly, we augment the actions to generate virtual trajectories in order to train the network to "see" more flexible experiences with cycle consistency constraint.
Our method contains three main components which we describe below.

\textbf{Dynamics Model for Prediction of Future States.}
A one-step Markov transition $(\mathbf{s}_t, \mathbf{a}_t, \mathbf{s}_{t+1})$ contains a current state $\mathbf{s}_t \in \mathcal{S}$, an action $\mathbf{a}_t \in \mathcal{A}$, and the next state $\mathbf{s}_{t+1} \in \mathcal{S}$. The transition model determines the next state $\mathbf{s}_{t+1}$ given the current state-action pair $(\mathbf{s}_t, \mathbf{a}_t)$.

Considering the ability to model what will happen next is important for RL tasks, many works train agents to be predictive of the future states to learn good feature representations~\cite{schwarzer2021dataefficient,Shelhamer2017LossII,guo2020bootstrap,lee2020slac}. 
Following SPR \cite{schwarzer2021dataefficient}, we introduce a dynamics model (DM) $h(\cdot, \cdot)$ to predict the transition dynamics  $(\mathbf{z}_t,\mathbf{a}_t)\rightarrow \mathbf{z}_{t+1}$ in the latent
feature space, where $\mathbf{z}_t = f(\mathbf{s}_t)$ is encoded by the feature encoder $f(\cdot)$ of the current input video clip $\mathbf{s}_t$.
As illustrated in Figure~\ref{fig:framework}(b), based on the current input state $\mathbf{z}_t$ and a sequence of actions $\mathbf{a}_{t:t+K-1}$, we obtain a sequence of $K$ predictions $\hat{\mathbf{z}}_{t+1:t+K}$ of the future state representations using the action-conditioned transition model (\ieno, DM) $h(\cdot, \cdot)$ by computing the next state iteratively as
\begin{equation}
    \hat{\mathbf{z}}_{t+k+1}=\begin{cases}
    h(\mathbf{z}_{t+k},\mathbf{a}_{t+k}) & \text{ if } k=0 \\ 
    h(\hat{\mathbf{z}}_{t+k},\mathbf{a}_{t+k}) & \text{ if } k=1,2,\cdots,K-1. 
    \end{cases}
\end{equation}
We train the DM with the supervision of the future state representations obtained from the recorded real trajectory (\ieno, from the recorded future video clip). Following SPR \cite{schwarzer2021dataefficient}, we compute the 
prediction loss by summing over difference (error) between the predicted 
representations $\hat{\mathbf{z}}_{t+k}$ and observed representations $\tilde{\mathbf{z}}_{t+k}$ at timesteps $t+k$ for $1 \leq k \leq K$ measured in a "projection" metric space as:
\begin{equation}
    \mathcal{L}_{pred}=  \sum_{k=1}^{K}{d (\hat{\mathbf{z}}_{t+k}, \tilde{\mathbf{z}}_{t+k})},
    \label{equ:pred_loss}
\end{equation}
where $d$ denotes the distance metric in a "projection" space~\cite{schwarzer2021dataefficient} (see Appendix \ref{subsec:network} for more details).

This module has two roles in our framework. (i) The future prediction helps to learn good feature state representation, which enables the scheme SPR~\cite{schwarzer2021dataefficient} that we use as our strong baseline. (ii) It paves the way for our introduction of cycle-consistent virtual trajectories for improving data efficiency.

\textbf{Backward Dynamics Model for Prediction of Previous States}:
Backward transition model intends to determine the previous state $\mathbf{s}_t$ given the next state $\mathbf{s}_{t+1}$ and the causal action $\mathbf{a}_{t}$. We introduce a backward dynamics model (BDM) $b(\cdot,\cdot)$ to predict the backward transition dynamics $(\mathbf{z}_{t+1},\mathbf{a}_t)\rightarrow \mathbf{z}_{t}$ in the latent feature space. 

In previous works~\cite{goyal2019recall,edwards2018forward, nair2020trass}, backward induction has been exploited to predict the preceding states that terminate at a given high-reward state, where these traces of (state, action) pairs are used to improve policy learning. Their purpose is to emphasize the training on high-reward states and the probable trajectories leading to them to alleviate the problem of lack of high reward states for policy learning. In contrast, we introduce a BDM which predicts previous states (to have a backward trajectory) in order to build a cycle/loop with the forward trajectory to enforce the consistency constraint for boosting feature representation learning. 

As illustrated in Figure~\ref{fig:framework}(c), based on the hidden state $\hat{\mathbf{z}}_{t+K}$ and a sequence of actions $\check{\mathbf{a}}_{t+K-1:t}$, we obtain a sequence of $K$ predictions $\mathbf{z}'_{t+K-1:t}$ of the previous state representations using the BDM $b(\cdot, \cdot)$ by computing the previous state iteratively as
\begin{equation}
    \mathbf{z}'_{t+k-1}=\begin{cases}
    b(\hat{\mathbf{z}}_{t+k},\check{\mathbf{a}}_{t+k-1}) & \text{ if } k=K \\ 
    b(\mathbf{z}'_{t+k},\check{\mathbf{a}}_{t+k-1}) & \text{ if } k=K-1,K-2,\cdots,1. 
    \end{cases}
\end{equation}

\textbf{Action Augmentation and Cycle Consistency Constraint.}
Given the DM, BDM, a current state, and a sequence of actions, we can easily generate a forward trajectory and a corresponding backward trajectory which forms a loop/cycle/forward-backward trajectory. 
As we know, for an encoder which is capable of encoding observations to suitable feature representations, the feature representations of the start state $\mathbf{z}_t$ and the end state $\mathbf{z}'_t$ of a forward-backward trajectory should in general be similar/consistent, given a reasonable DM and BDM. 

Therefore, as illustrated in Figure \ref{fig:framework-c}, we enforce a consistency constraint between the start state $\mathbf{z}_t$ and the end state $\mathbf{z}'_t$ to regularize the feature representation learning. In this way, by augmenting actions (generating/sampling virtual actions), we can obtain abundant virtual cycle-consistent trajectories for training. Note that in the training, we do not need supervision of the states from real trajectories. 

Here, we mathematically define the cycle-consistent feature representation in a forward-backward trajectory as below. 
\newtheorem{definition}{Definition}
\begin{definition}
  Given a (forward) dynamics model $h$ and a backward dynamics model $b$, cycle-consistent feature representation $\mathbf{z}_t$ in a forward-backward trajectory ${\tau}^{c}$ is a representation of the current state that meets the following condition when experiencing any sequence of $K$ actions $\{\mathbf{a}_{t}, \mathbf{a}_{t+1}, \dots, \mathbf{a}_{t+K-1}\}$ sampled from an action space $\mathcal{A}$:
    \begin{align*}
        \mathbb{E}_{{\tau}^{c}}[\mathit{d}_\mathcal{M}(\mathbf{z}'_t,\mathbf{z}_t)]=0,
    \end{align*}
where $\mathit{d}_\mathcal{M}$ is a distance metric on space $\mathcal{M}$ and $\mathbf{z}'_t$ is the prediction of $\mathbf{z}_t$ after experiencing a sequence of actions in forward prediction and backward prediction as

\begin{align*}
\begin{split}
forward
:~~ &  \hat{\mathbf{z}}_{t} = \mathbf{z}_{t}, ~\hat{\mathbf{z}}_{t+k+1} = h(\hat{\mathbf{z}}_{t+k}, \mathbf{a}_{t+k}),   ~~\text{for}~ k=0,1,\cdots,K-1,\\
backward:~~ &  \mathbf{z}'_{t+K}=\hat{\mathbf{z}}_{t+K}, ~\mathbf{z}'_{t+k} = b(\mathbf{z}'_{t+k+1}, \mathbf{a}_{t+k}),  ~~\text{for}~ k=K-1,K-2,\cdots,0.
\end{split} 
\end{align*}
\end{definition}

Given the state $\mathbf{z}_t$ encoded from the current input $\mathbf{s}_t$ of time $t$, we \tcr{randomly} sample $M$ sets of actions \tcr{in the action space $\mathcal{A}$}.
We calculate the cycle consistency loss as:
\begin{equation}
    \mathcal{L}_{cyc}=\frac{1}{M}\sum_{m=1}^{M}{\mathit{d}_\mathcal{M}(\mathbf{z}'_t, \mathbf{z}_{t})}.
    \label{equ:cycle_loss}
\end{equation}
We describe the alternative distance metrics $\mathit{d}_\mathcal{M}$ on space $\mathcal{M}$ and study the influence on performance in \tcb{Section \ref{ablation}}.

\emph{Discussion}\label{discussion}: In our scheme, similar to \cite{goyal2019recall,edwards2018forward, nair2020trass}, we model the backward dynamics using a BDM. 
This is basically feasible for many real-world applications, \egno, robotic control, and many games. 
Consider a robotic arm: given a current state (\egno, spatial position and rotation) and the previous action (\egno, quaternions or six-degree-of-freedom parameters), one can deduce the previous state without much effort.
This holds in most games such as chess or Atari.
\tcr{For} some cases where there are many-to-one transition (\ieno, different states with the same action may be transited into the same state), BDM \tcr{may} learn the most probable state or a mean state to minimize the prediction errors \tcr{through} the powerful fitting capacity of \tcr{neural networks}. \tcr{More discussion can be found in Appendix \ref{discussion}}.

\textbf{Overall Training Objective.}
The overall training objective of our method is as below:
\begin{equation}
    \mathcal{L}_{total}=\mathcal{L}_{rl}+\lambda_{pred}\mathcal{L}_{pred}+\lambda_{cyc}\mathcal{L}_{cyc},
    \label{equ:total_loss}
\end{equation}
where $\mathcal{L}_{rl}$, $\mathcal{L}_{pred}$, and $\mathcal{L}_{cyc}$ denote the RL loss (please refer to Rainbow \cite{hessel2018rainbow} for discrete control games, SAC \cite{haarnoja2018soft} for continuous control games), prediction loss (see Eq.~(\ref{equ:pred_loss})), and cycle consistency loss (see Eq.~(\ref{equ:cycle_loss})), respectively. $\lambda_{pred}$ and $\lambda_{cyc}$ are the hyperparameters for balancing the contributions of different losses.

\section{Experiments}
We introduce the experimental setup including environments, evaluation, and implementation details. We conduct extensive ablation studies to demonstrate and analyze the effectiveness of our designs.

\subsection{Setup}
\label{subsec:setup}
\textbf{Environments and Evaluation.} 
We evaluate our method on the commonly used discrete control benchmark of Atari~\cite{bellemare2013arcade}, and the  continuous control benchmark of DMControl~\cite{tassa2018deepmind}. 
Following \cite{laskin2020curl,yarats2021image}, we measure the performance of different methods at 100k \emph{interaction steps} (400k environment steps with action repeat of 4) on Atari (also refer to as \textbf{Atari-100k}), and at 100k and 500k \emph{environment steps} on DMControl (refer to as \textbf{DMControl-100k} or \textbf{DMC-100k}, \textbf{DMControl-500k} or \textbf{DMC-500k}), respectively. In general, using Atari-100k on 26 selected games~\cite{Kaiser2020Model,van2019der,kielak2020recent,laskin2020curl} and DMControl-100k~\cite{laskin2020curl,yarats2021image} has been a common practice for investigating data efficiency. 

For Atari-100k, we measure the performance by score, and the median human-normalized score (\ieno, \emph{median HNS}) of the 26 games. Human-normalized score on a game is calculated by $\frac{S_A-S_R}{S_H-S_R}$, where $S_A$ is the agent score, $S_R$ is the score of random play, and $S_H$ is the expert human score. For DMControl, the maximum possible score for every environment is $1000$ \cite{tassa2018deepmind}. Following \cite{laskin2020curl,yarats2019improving,hafner2019learning,yarats2021image,liu2021returnbased}, we evaluate models on the six commonly used DMControl environments. Additionally, we use the \emph{median score} on them to reflect the overall performance.

\textbf{Implementation Details.} 
For the discrete control benchmark of \textbf{Atari}, we use SPR~\cite{schwarzer2021dataefficient} as our strong baseline (dubbed \emph{Baseline}) and build our method on top of SPR by augmenting cycle-consistent virtual trajectories for better representation learning (see Figure~\ref{fig:framework-c}). 
For the backward dynamics model, we use the same architecture as that of the dynamics model.
We set the number of prediction steps $K$ to $9$ by default. We simply set the number of action sets, \ieno, the number of virtual trajectories $M$ to $2 |\mathcal{A}|$, which is proportional to the size of action space  $|\mathcal{A}|$ in that 
Atari game. 
\tcr{To generate an action sequence, we randomly sample an action from the discrete action space at each step.}
\emph{We report the results of SPR~\cite{schwarzer2021dataefficient} by re-running their source code in all Tables except for Table~\ref{table:atari_compare} (results in which are copied from their paper, being higher than our reproduced results).}

For the continuous control benchmark of \textbf{DMControl}, considering the SPR is originally designed only for discrete control, we build a SPR-like scheme SPR\bmark as our baseline (dubbed \emph{Baseline}) for continuous control games. Particularly, we use the encoder and policy networks of CURL \cite{laskin2020curl} as the basic networks. Following SPR~\cite{schwarzer2021dataefficient}, we remove the contrastive loss in CURL and introduce BYOL~\cite{grill2020byol} heads to build SPR-like baseline scheme. 
We use the network architecture similar to the dynamics model in DBC \cite{zhang2021learning} to build the dynamics model in SPR\bmark~and the backward dynamics model in our \ourname.
We follow the \tcb{training settings} in CURL except the batch size (\tcb{reduced} from 512 to 128 to save memory cost) and learning rate. We set $K$ to $6$\tcb{, and} set $M$ to a fixed number 10 with actions 
\tcb{randomly sampled from the uniform distribution of the continuous action space.}

We set $\lambda_{pred}=1$ and $\lambda_{cyc}=1$. For $\mathit{d}_\mathcal{M}$, we use the distance metric as in SPR \cite{schwarzer2021dataefficient}. 
More implementation details can be found in Appendix \ref{details}. All our models are implemented via PyTorch~\cite{paszke2019pytorch}.

\subsection{Performance Comparison with State-of-the-Arts}
\begin{table*}[h]
    \footnotesize 
    \centering
    \caption{Scores achieved by different methods on Atari-100k. We also report median HNS. We run our \ourname~with 15 random seeds given that this benchmark is susceptible to high variance across multiple runs. Note that here we report the results of SPR~\cite{schwarzer2021dataefficient} copied from their paper (\ieno, 41.5\%), which is much higher than our reproduced results using their released source code (\ieno, 37.1\%).}   
    \label{table:atari_compare}
    \scalebox{0.77}{
        \begin{tabular}{l c c c c c c c c c c}
            \toprule
            \textbf{Game} & \textbf{Human} & \textbf{Random} & \textbf{SimPLe}\cite{Kaiser2020Model} & \textbf{DER}\cite{van2019der} & \textbf{OTR}\cite{kielak2020recent} & \textbf{CURL}\cite{laskin2020curl} & \textbf{DrQ}\cite{yarats2021image} & \textbf{SPR}\cite{schwarzer2021dataefficient} & \textbf{\ourname}\\
            \midrule
            Alien           & 7127.7    & 227.8     & 616.9     & 739.9     & 824.7     & 558.2     & 771.2     & 801.5     & \textbf{947.8}\\
            Amidar          & 1719.5    & 5.8       & 88.0      & \textbf{188.6}     & 82.8      & 142.1     & 102.8     & 176.3     & 165.3 \\
            Assault         & 742.0     & 222.4     & 527.2     & 431.2     & 351.9     & 600.6     & 452.4     & 571.0     & \textbf{702.3}\\
            Asterix         & 8503.3    & 210.0     & \textbf{1128.3}    & 470.8     & 628.5     & 734.5     & 603.5     & 977.8     & 933.3\\
            Bank Heist      & 753.1     & 14.2      & 34.2      & 51.0      & 182.1     & 131.6     & 168.9     & \textbf{380.9}     & 245.9\\
            Battle Zone     & 37187.5   & 2360.0    & 5184.4    & 10124.6   & 4060.6    & 14870.0   & 12954.0   & \textbf{16651.0}   & 13260.0\\
            Boxing          & 12.1      & 0.1       & 9.1       & 0.2       & 2.5       & 1.2       & 6.0       & 35.8      & \textbf{38.3} \\
            Breakout        & 30.5      & 1.7       & 16.4      & 1.9       & 9.8      & 4.9       & 16.1      & 17.1      & \textbf{20.6}\\
            Chopper Command & 7387.8    & 811.0     & \textbf{1246.9}    & 861.8     & 1033.3   & 1058.5    & 780.3     & 974.8     & 922.4\\
            Crazy Climber   & 35829.4   & 10780.5   & \textbf{62583.6}   & 16185.3   & 21327.8   & 12146.5   & 20516.5   & 42923.6   & 23176.7\\
            Demon Attack    & 1971.0    & 152.1     & 208.1     & 508.0     & 711.8     & 817.6     & 1113.4    & 545.2     & \textbf{1131.7} \\
            Freeway         & 29.6      & 0.0       & 20.3      & \textbf{27.9}      & 25.0      & 26.7      & 9.8       & 24.4      & 16.1 \\
            Frostbite       & 4334.7    & 65.2      & 254.7     & 866.8     & 231.6     & 1181.3    & 331.1     & 1821.5    & \textbf{1984.7}\\
            Gopher          & 2412.5    & 257.6     & 771.0     & 349.5     & \textbf{778.0}     & 669.3     & 636.3     & 715.2     & 684.3\\
            Hero            & 30826.4   & 1027.0    & 2656.6    & 6857.0    & 6458.8    & 6279.3    & 3736.3    & 7019.2    & \textbf{8597.5} \\
            Jamesbond       & 302.8     & 29.0      & 125.3     & 301.6     & 112.3     & \textbf{471.0}     & 236.0     & 365.4     & 394.7\\
            Kangaroo        & 3035.0    & 52.0      & 323.1     & 779.3     & 605.4     & 872.5     & 940.6     & \textbf{3276.4}    & 2384.7\\
            Krull           & 2665.5    & 1598.0    & \textbf{4539.9}    & 2851.5    & 3277.9    & 4229.6    & 4018.1    & 3688.9    & 3880.7\\
            Kung Fu Master  & 22736.3   & 258.5     & \textbf{17257.2}   & 14346.1   & 5722.2    & 14307.8   & 9111.0    & 13192.7   & 14259.0\\
            Ms Pacman       & 6951.6    & 307.3     & \textbf{1480.0}    & 1204.1    & 941.9     & 1465.5    & 960.5     & 1313.2    & 1335.4\\
            Pong            & 14.6      & -20.7     & \textbf{12.8}      & -19.3     & 1.3       & -16.5     & -8.5      & -5.9      & -3.0 \\
            Private Eye     & 69571.3   & 24.9      & 58.3      & 97.8      & 100.0     & \textbf{218.4}     & -13.6     & 124.0     & 93.9\\
            Qbert           & 13455.0   & 163.9     & 1288.8    & 1152.9    & 509.3     & 1042.4    & 854.4     & 669.1     & \textbf{3620.1}\\
            Road Runner     & 7845.0    & 11.5      & 5640.6    & 9600.0    & 2696.7    & 5661.0    & 8895.1    & \textbf{14220.5}   & 13534.0\\
            Seaquest        & 42054.7   & 68.4      & \textbf{683.3}     & 354.1     & 286.9    & 384.5     & 301.2     & 583.1     & 527.7 \\
            Up N Down       & 11693.2   & 533.4     & 3350.3    & 2877.4    & 2847.6    & 2955.2    & 3180.8    & \textbf{28138.5}   & 10225.2\\
            \midrule
            Median HNS ($\%$)      & 100   & 0         & 14.4  & 16.1  & 20.4  & 17.5  & 26.8  & 41.5  & \textbf{47.2}\\
            \bottomrule
        \end{tabular}
    }
    \vspace{-3pt}   
\end{table*}

\textbf{Comparison on Atari.}
On Atari-100k, Table \ref{table:atari_compare} shows the comparisons with the state-of-the-art methods.
We also report the results of random play (Random) and expert human play (Human) (copied from~\cite{wang2016dueling}). \ourname~achieves a median HNS of 47.2\%, significantly outperforming all previous methods. \ourname~surpasses the baseline SPR~\cite{schwarzer2021dataefficient}(with a median HNS of 41.5\% reported in their paper) by \textbf{5.7\%}. We have re-run the released source code of SPR with 15 random seeds and obtain a median HNS of 37.1\%, which suggests that our improvement over SPR is actually \textbf{10.1\%}. 

\begin{table*}[t]
    \vspace{0pt}
    \footnotesize 
    \centering
    \caption{Scores (mean and standard deviation) achieved by different methods on the DMControl-100k and DMControl-500k. We run our \ourname~with 10 random seeds. Note that SPR \cite{schwarzer2021dataefficient} is originally designed only for discrete control. \tcr{For the continuous-control environments, we extend SPR to a new version named SPR\bmark.}
    }
    \label{table:dmc_compare}
    \scalebox{0.77}{
        \begin{tabular}{l c c c c c c c c}
            \toprule
            \textbf{100k Step Scores} & \textbf{PlaNet}\cite{hafner2019learning} & \textbf{Dreamer}\cite{Hafner2020Dream} & \textbf{SAC+AE}\cite{yarats2019improving} & \textbf{SLAC}\cite{lee2020slac}  & \textbf{CURL}\cite{laskin2020curl} & \textbf{DrQ} \cite{yarats2021image} & \textbf{SPR\bmark}\cite{schwarzer2021dataefficient} & \textbf{\ourname} \\ \hline
            \specialrule{0em}{1.5pt}{1pt}   
            Finger, spin & 136 $\pm$ 216  & 341 $\pm$ 70   & 740 $\pm$ 64    & 693 $\pm$ 141 & 767 $\pm$ 56  & 901 $\pm$ 104  & 868 $\pm$ 143  & \textbf{915 $\pm$ 49} \\
            Cartpole, swingup & 297 $\pm$ 39   & 326 $\pm$ 27   & 311 $\pm$ 11    & -             & 582 $\pm$ 146 &  759 $\pm$ 92 & 799 $\pm$ 42   & \textbf{816 $\pm$ 36} \\
            Reacher, easy & 20 $\pm$ 50    & 314 $\pm$ 155  & 274 $\pm$ 14    & -             & 538 $\pm$ 233  & 601 $\pm$ 213 & 638 $\pm$ 269   & \textbf{785 $\pm$ 142} \\
            Cheetah, run & 138 $\pm$ 88   & 235 $\pm$ 137  & 267 $\pm$ 24    & 319 $\pm$ 56  & 299 $\pm$ 48  & 344 $\pm$ 67  & 467 $\pm$ 36   & \textbf{474 $\pm$ 50} \\
            Walker, walk  & 224 $\pm$ 48   & 277 $\pm$ 12   & 394 $\pm$ 22    & 361 $\pm$ 73  & 403 $\pm$ 24  & \textbf{612 $\pm$ 164}  & 398 $\pm$ 165   & 460 $\pm$ 173 \\
            Ball in cup, catch & 0 $\pm$ 0      & 246 $\pm$ 174  & 391 $\pm$ 82    & 512 $\pm$ 110 & 769 $\pm$ 43  & 913 $\pm$ 53  & 861 $\pm$ 233  & \textbf{926 $\pm$ 31} \\ 
            \midrule
            Median Score & 137.0 & 295.5 & 351.0 & 436.5 & 560.0 & 685.5 & 719.0 & \textbf{800.5} \\ 
            \midrule
            \textbf{500k Step Scores}  & & & & & & & & \\ \midrule
            Finger, spin & 561 $\pm$ 284  & 796 $\pm$ 183  & 884 $\pm$ 128   & 673 $\pm$ 92  & 926 $\pm$ 45 & 938 $\pm$ 103 & 924 $\pm$ 132 & \textbf{963 $\pm$ 40} \\
            Cartpole, swingup & 475 $\pm$ 71   & 762 $\pm$ 27   & 735 $\pm$ 63    & -             & 841 $\pm$ 45& 868 $\pm$ 10 & \textbf{870 $\pm$ 12}  & 865 $\pm$ 11 \\
            Reacher, easy & 210 $\pm$ 390  & 793 $\pm$ 164  & 627 $\pm$ 58    & -             & 929 $\pm$ 44 & 942 $\pm$ 71 & 925 $\pm$ 79 &\textbf{942 $\pm$ 66} \\
            Cheetah, run & 305 $\pm$ 131  & 570 $\pm$ 253  & 550 $\pm$ 34    & 640 $\pm$ 19  & 518 $\pm$ 28  & 660 $\pm$ 96 & 716 $\pm$ 47 &\textbf{719 $\pm$ 51}  \\
            Walker, walk  & 351 $\pm$ 58   & 897 $\pm$ 49   & 847 $\pm$ 48    & 842 $\pm$ 51  & 902 $\pm$ 43  & 921 $\pm$ 45 & 916 $\pm$ 75 &\textbf{928 $\pm$ 30}\\
            Ball in cup, catch  & 460 $\pm$ 380  & 879 $\pm$ 87   & 794 $\pm$ 58    & 852 $\pm$ 71 & 959 $\pm$ 27 & 963 $\pm$ 9 & 963 $\pm$ 8 &\textbf{967 $\pm$ 5} \\ 
            \midrule
            Median Score & 405.5 & 794.5 & 764.5 & 757.5 & 914.0 & 929.5 & 920.0 & \textbf{935.0} \\
            \bottomrule
        \end{tabular}
    }
    \vspace{-5pt}
\end{table*}

\textbf{Comparison on DMControl.}
For each environment in DMControl, we run our \ourname~with 10 random seeds to report the results. 
Table \ref{table:dmc_compare} shows the comparisons with the state-of-the-art methods.
Our method performs the best for the majority (\textbf{5} out of \textbf{6}) of the environments on both DMControl-100k and DMControl-500k. 
(i) On DMControl-100k which is in low data regime, our method achieves the highest median score of 800.5, which is about \tcr{\textbf{11.3\%} higher than SPR\bmark, 16.7\%} higher than DrQ~\cite{yarats2021image} and 42.9\% higher than CURL~\cite{laskin2020curl}. (ii) On DMControl-500k, our method achieves a median score of 935.0, which approaches the perfect score of 1000 and outperforms all other methods. 
Therefore, our method achieves superior performance in both data-efficiency and asymptotic performance. 

\subsection{Ablation Studies} \label{ablation}
We use the median HNS and median score to measure the overall performance on Atari and DMControl, respectively.
We run each game in Atari with 15 random seeds. To save computational resource, we run each environment in DMControl with 5 random seeds (instead of 10 as in Table~\ref{table:dmc_compare}).


\begin{wraptable}{r}{7.cm}
\vspace{-4mm}
    \centering
    \caption{Effectiveness of \ourname~on top of \emph{Baseline}, which is SPR~\cite{schwarzer2021dataefficient} for discrete control on Atari, and SPR\bmark for continual control on DMControl. "w/o Pred" denotes disabling future prediction in \emph{Baseline}. \emph{Baseline+BDM} denotes the scheme that a BDM is incorporated into \emph{Baseline}. }
    \label{table:ablation_loss}
    \scalebox{0.85}{
        \begin{tabular}{l c c }
            \toprule
            \textbf{Model} & $\textbf{Atari-100k}$ & $\textbf{DMControl-100k}$\\
            \midrule
            Baseline w/o~Pred  & 33.4 & 680.0 \\
            Baseline     & 37.1 & 728.0 \\
            Baseline+BDM & 38.4 & 741.0  \\
            \midrule
            \ourname      & \textbf{47.2} & \textbf{797.0} \\
            \bottomrule
        \end{tabular}
    }
\end{wraptable}

\textbf{Effectiveness of \ourname.} As described in Section~\ref{subsec:setup}, we take SPR~\cite{schwarzer2021dataefficient} as our baseline (\ieno, \emph{Baseline}) on discrete control benchmark Atari, and SPR\bmark on continuous control benchmark DMControl. 
On top of \emph{Baseline}, we validate the effectiveness of our \ourname~which augments cycle-consistent virtual trajectories for improving data efficiency. Table \ref{table:ablation_loss} shows the comparisons. We can see that \emph{\ourname}~achieves a median HNS of 47.2\%, which outperforms \emph{Baseline} by \textbf{10.1\%} on Atari-100k. On DMControl-100, \emph{\ourname} improves \emph{Baseline} from 728.0 to 797.0 in terms of median score (\ieno, a relative gain of 9.5\%). As a comparison, \emph{Baseline} outperforms \emph{Baseline w/o~Pred} by 3.7\% on Atari-100k, where "Pred" denotes the prediction of future state in SPR / SPR\bmark (\ieno, the contribution of SPR~\cite{schwarzer2021dataefficient}). The large gains of our \ourname~over \emph{Baseline} demonstrate the effectiveness of our \ourname~in boosting feature representation learning. 
In addition, to further benchmark \ourname's data efficiency, we compare the testing performance in every 5k environment steps at the first 100k on DMControl, where the result curves in Appendix \ref{subsec:more_ablations} show that our \ourname~consistently outperforms \emph{Baseline}.

One may wonder whether the major performance gain of our \ourname~is attributed to the introduction of backward dynamics model (BDM) or by our augmentation of virtual trajectories. When we disable the augmentation of virtual trajectories, our scheme degrades to \emph{Baseline+BDM}, where a BDM is incorporated into the baseline SPR (or SPR\bmark) and only the real trajectories go through the BDM. In Table \ref{table:ablation_loss}, we can see that introducing BDM does not improve the performance obviously and our augmentation of cycle-consistent virtual trajectories for regularizing feature representation learning is the key for the success.


\begin{wraptable}{r}{7.cm}
\vspace{-4mm}
    \centering
    \caption{Influence of prediction steps $K$ for our \ourname~and the baseline scheme SPR/SPR\bmark.}
    \label{table:ablation_K}
    \scalebox{0.67}{
        \begin{tabular}{l l c c c c c}
            \toprule
            \textbf{Benchmark} & \textbf{Model} & \textbf{$K$=0} & \textbf{$K$=3} & \textbf{$K$=6} & \textbf{$K$=9} & \textbf{$K$=12} \\
            \midrule
            \multirow{2}{*}{Atari-100k}  & SPR & 33.4 & 33.9 & 35.2 & 37.1 & 34.9\\
            & \ourname & 33.4 & 34.8 & 39.2 & \textbf{47.2} & 43.1 \\
            \midrule
            \multirow{2}{*}{DMC-100k}  & SPR\bmark & 664.0 & 725.0 & 723.0 & 728.0 & 721.5\\
            & \ourname & 664.0 & 775.5 & \textbf{797.0} & 795.0 & 794.5 \\
            \bottomrule
        \end{tabular}
    }
    \vspace{-4pt}   
\end{wraptable}
\textbf{Influence of Prediction Steps $K$.}
We study the influence of $K$ for both our \ourname~and the baseline scheme SPR/SPR\bmark.
Table~\ref{table:ablation_K} shows the performance. When $K=0$, both schemes degrade to \emph{Baseline w/o Pred} (where future prediction is disabled in SPR/SPR\bmark). We have the following observations/conclusions. (i) Given the same number of prediction steps $K$ (beside 0), our \ourname~consistently outperforms the baseline scheme SPR/SPR\bmark on both benchmarks Atari-100k/MDControl-100k. (ii) Our \ourname~achieves the best performance at $K=9$ on Atari and $K=6$ on DMControl, which outperforms the baseline at the same $K$ by 10.1\% and 9.5\% (relative gain) on Atari and DMControl, respectively. Note that the performance of SPR~\cite{schwarzer2021dataefficient} obtained using their source code at $K$=5 (note $K$=5 is used in SPR paper) is 36.1\% (which is 41.5\% in their paper) on Atari-100k. (iii) In SPR~\cite{schwarzer2021dataefficient}/SPR\bmark, a too small number of prediction steps cannot make the feature representation sufficiently predictive of the future while a too large number of prediction steps may make the RL loss contributes less to the feature representation learning (where a more elaborately designed weight $\lambda_{pred}$ is needed). Our \ourname~follows similar trends.

\begin{wraptable}{r}{7.cm}
\vspace{-4mm}
\centering
\caption{Impact of the augmentation of cycle-consistent virtual trajectories on feature representation learning. \emph{\ourname-ND} denotes that we do not use the cycle consistency loss over virtual trajectories to update the dynamic model.}
\scalebox{0.88}{
        \begin{tabular}{l c c}
            \toprule
            \textbf{Model} & $\textbf{Atari-100k}$ & $\textbf{DMControl-100k}$ \\
            \midrule
            Baseline       &  37.1 & 723.0 \\
            \ourname-ND & 44.0  & 777.5 \\
            \ourname & \textbf{47.2} & \textbf{797.0} \\
            \bottomrule
        \end{tabular}} \label{table:ablation_DM}
    \vspace{-3.0pt}   
\end{wraptable}
\textbf{What does Augmenting Cycle-Consistent Virtual Trajectories Help?}
We propose the augmentation of cycle-consistent virtual trajectories in order to boost the feature representation learning of RL for improving data efficiency. In the training, the cycle consistency loss $L_{cyc}$ over the virtual trajectories would optimize the parameters of the encoder, DM and \tcr{BDM}. One may wonder what the gain is mainly coming from. Is it because the DM is more powerful/accurate that enables better prediction of future states? Or is it because the encoder becomes more powerful to provide better feature representation? We validate this by letting the cycle consistency loss $L_{cyc}$ not update DM, where DM is only optimized by prediction loss $L_{pred}$ as in SPR. We denote this scheme as \emph{\ourname-ND}. Table \ref{table:ablation_DM} shows that we obtain a gain of 6.9\% in \emph{\ourname-ND} from the regularization of $L_{cyc}$ \emph{on the encoder} and a gain of 10.1\% in \emph{\ourname}~from the regularization \emph{on both the encoder and DM} on Atari. Similar trend is observed on DMControl. This implies that the augmentation of cycle-consistent virtual trajectories is helpful to DM training but the main gain is brought by its regularization on the feature representation learning of the encoder.

\tcr{
\textbf{Influence of Distance Metric $\mathit{d}_\mathcal{M}$ on Space $\mathcal{M}$.}
For the distance metric $d_{\mathcal{M}}$ in space $\mathcal{M}$, we compare cosine distance on the latent feature space $\mathcal{M}_{latent}$, \ieno, $\mathit{d}_\mathcal{M}(\mathbf{z}'_t,\tilde{\mathbf{z}}_t) = 2-2\frac{\mathbf{z}'_t}{\|\mathbf{z}'_t\|} \frac{\tilde{\mathbf{z}}_t}{\|\tilde{\mathbf{z}}_t\|}$
 and on the "projection" space $\mathcal{M}_{proj}$ as in SPR~\cite{schwarzer2021dataefficient} (see Appendix \ref{subsec:network} for more details). 
We compare the influence of feature space for calculating cycle consistency loss and show the results in Table \ref{table:ablation_metric}. 
On the Atari benchmark, our \ourname~with distance metric on space $\mathcal{M}_{latent}$ and with distance metric on space $\mathcal{M}_{proj}$ significantly outperforms \emph{Baseline} by 7.7\% and 10.1\%, respectively. This demonstrates the effectiveness of our key idea of \tcb{exploiting virtual trajectories} for effective representation learning. $\mathcal{M}_{proj}$ performs 2.4\% better than $\mathcal{M}_{latent}$. That maybe because for \ourname~and \emph{Baseline} for Atari, latent feature $\mathbf{z}_t$ (which corresponds to a $64\times7\times7$ feature map) preserves more spatial information than projected feature, where the former is less robust to be matched across two augmented \tcb{observations} due to spatial misalignment. On the DMControl benchmark, our \ourname~with distance metric on space $\mathcal{M}_{latent}$ and with distance metric on space $\mathcal{M}_{proj}$ significantly outperforms \emph{Baseline} by 70.5 and 69.0 \tcb{in terms of median score}, respectively. The performance of $\mathcal{M}_{latent}$ and $\mathcal{M}_{proj}$ are similar. Note that the latent feature $\mathbf{z}_t$ of \ourname~or \emph{Baseline} (built based on CURL) corresponds to a feature vector which is obtained after a fully connected layer in the backbone network, which does not face the spatial misalignment problem caused by the augmentation. \tcb{We use $\mathcal{M}_{proj}$ as the default metric space in this work.}
\begin{table*}[h]
    \centering
    \caption{Influence of distance metric space $\mathcal{M}$. $\mathcal{M}_{latent}$ and $\mathcal{M}_{proj}$ denote the use of the latent feature space and the "projection" space, respectively.}
    \label{table:ablation_metric}
    \scalebox{0.88}{
        \begin{tabular}{l c c c }
            \toprule
            \textbf{Model} & $\textbf{Atari-100k}$ & $\textbf{DMControl-100k}$\\
            \midrule
            Baseline & 37.1 & 728.0 \\
            \ourname($\mathcal{M}_{latent}$) & 44.8 & \textbf{798.5} \\
            \ourname($\mathcal{M}_{proj}$) & \textbf{47.2} & 797.0 \\
            \bottomrule
        \end{tabular}
    }
\end{table*}

}

\tcr{
\textbf{Influence of the Number of Virtual Trajectories $M$.}
Table \ref{table:ablation_M} shows the influence of the number of virtual trajectories $M$. We can observe that small $M$ (less generated virtual trajectories) is inferior to a suitable $M$. That may be because \tcb{too small $M$} cannot cover diverse experiences for feature representation learning. When $M$ is too large, it brings less additional benefit. That may be because a suitable number of trajectories is enough for regularizing the network training. We observe that the performance drops when $M$ is too large. That may be because a very large $M$ would increase the optimization difficulty in practice.
\begin{table*}[h]
    \centering
    \caption{Influence of the number of virtual trajectories $M$.}
    \label{table:ablation_M}
    \scalebox{0.88}{
        \begin{tabular}{l c c c c c}
            \toprule
            \textbf{Atari-100k} &  & & & &\\
            \midrule
            M           & 0 & $|\mathcal{A}|$ & $2|\mathcal{A}|$ & $3|\mathcal{A}|$ &\\
            Median HNS(\%)  & 37.1 & 39.5 & \textbf{47.2} & 42.5 &\\
            \midrule
            \textbf{DMControl-100k} &  & &  & &\\
            \midrule
            M           & 0 & 1 & 10 & 20 & 30 \\
            Median Score  & 723.0 & 770.5 & 797.0 & \textbf{806.0} & 792\\
            \bottomrule
        \end{tabular}
    }
\end{table*}

}

\section{Conclusion} \label{sec:conclusion}
With limited experience, deep RL suffers from data inefficiency.
In this work, we propose a new method, dubbed \ourname, which augments cycle-consistent virtual state-action trajectories to enhance the data efficiency for RL feature representation learning. \ourname~predicts future states based on the current state and a sequence of sampled actions and then predicts the previous states, which forms a trajectory cycle/loop. We enforce the trajectory to meet the cycle consistency constraint to regularize the feature representation learning. Experimental results on both the discrete control benchmark Atari and continuous control benchmark DMControl demonstrate the effectiveness of our method, where we achieve the state-of-the-art performance on both benchmarks.

\begin{ack}
This work was supported in part by the National Key Research and Development Program of China 2018AAA0101400 and NSFC under Grant U1908209, 61632001 and 62021001.
\end{ack}

\newpage
\bibliography{neurips_2021}
\bibliographystyle{neurips_2021}

\newpage
\section*{Appendix}
\appendix
\section{More Implementation Details} \label{details}
\subsection{Network Architecture}
\label{subsec:network}
\textbf{Network Architecture for Discrete Control Benchmark of Atari.}
For the discrete control benchmark of \textbf{Atari}, we use SPR~\cite{schwarzer2021dataefficient} as our strong baseline (dubbed \emph{Baseline}) and build our method on top of SPR by augmenting cycle-consistent virtual trajectories for better representation learning.

SPR~\cite{schwarzer2021dataefficient} has three main components: (online) encoder $f(\cdot)$, dynamics model (DM) $h(\cdot,\cdot)$, and policy learning (Q-learning) head $\pi(\cdot)$. The encoder consists of three convolutional layers with ReLU layer after each convolutional layer. The DM is composed of two convolutional layers with batch normalization \cite{ioffe2015batch} after the first convolutional layer and ReLU after the second convolutional layer. The Q-learning head is designed following Rainbow \cite{hessel2018rainbow}. 
Rather than predicting representations produced by the online encoder (by the DM), SPR computes target representations for future states using a target encoder $f_m$, whose parameters are an exponential moving average (EMA) of the online encoder parameters. 
To obtain the "projection" metric space $d$ (see Eq.~(ii) in the main manuscript) for future state prediction optimization, SPR uses online and target projection heads $g(\cdot)$ and $g_m(\cdot)$ to project online and target representations to a smaller latent space, and apply a prediction head $q(\cdot)$ to the online projections to predict the target projections.

For our \ourname, on top of SPR, we add a backward dynamics model (BDM) $b(\cdot,\cdot)$. For simplicity, we use the same network architecture as the DM. To calculate the cycle consistency loss for the feature representations (in a forward-backward trajectory) in a distance metric on space $\mathcal{M}$, we can simply use the cosine distance on the latent feature space, \ieno, $\mathit{d}_\mathcal{M}(\mathbf{z}'_t,\mathbf{z}_t) = 2-2\frac{\mathbf{z}'_t}{\|\mathbf{z}'_t\|} \frac{\mathbf{z}_t}{\|\mathbf{z}_t\|}$. As a design alternative, we can use the "projection" metric space as in SPR~\cite{schwarzer2021dataefficient} (discussed in the last paragraph) to calculate the cosine distance on the projection space, \ieno, 
$\mathit{d}_\mathcal{M}(\mathbf{z}'_t,\mathbf{z}_t) = 2-2\frac{q(g(\mathbf{z}'_t)}{\|q(g(\mathbf{z}'_t))\|} \frac{g_m(\mathbf{z}_t)}{\|g_m(\mathbf{z}_t)\|}$. 
In our implementation, we could directly use $\mathbf{z}_t$ (the start state of the virtual trajectory) as the target feature representation. Motivated by SPR, for each trajectory, we use the feature representation $\tilde{\mathbf{z}}_t$ of a stochastic augmentation $\tilde{\mathbf{s}}_t$ of the current video clip (observation) $\mathbf{s}_t$, as the target feature representation. Then, $\mathit{d}_\mathcal{M}(\mathbf{z}'_t,\tilde{\mathbf{z}}_t)$ is the actual distance metric.

\textbf{Network Architecture for Continuous Control Benchmark of DMControl.}
For the continuous control benchmark of \textbf{DMControl}, considering the SPR is originally designed only for discrete control, we build a SPR-like scheme SPR\bmark as our baseline (dubbed \emph{Baseline}) for continuous control games. 
Particularly, we use the encoder and policy networks of CURL \cite{laskin2020curl} as the basic networks. Following SPR~\cite{schwarzer2021dataefficient}, we remove the contrastive loss in CURL and introduce BYOL~\cite{grill2020byol} heads to build SPR-like baseline scheme. We use the network architecture similar to the dynamics model in DBC \cite{zhang2021learning} to build the dynamics model (DM) in SPR\bmark, where the DM consists of two fully connected layers with an LN (layer normalization) layer and a ReLU after the first fully connected layer.
The encoder has four convolutional layers (with a ReLU after each), followed by a fully connected layer, an LN layer~\cite{ba2016layer}, and a hyperbolic tangent (tanh) activation. Similar to the design in SPR, we have a projection head $g(\cdot)$, a prediction head $q(\cdot)$ for the (online) encoder, and a momentum encoder ${f}_m(\cdot)$ and a momentum projection head ${g}_m(\cdot)$. The projection head and prediction head are both built by two fully connected layers (with a ReLU layer after the first) of 512 hidden units for each.

For our \ourname, we add a backward dynamics model (BDM) $b(\cdot,\cdot)$ which has the same architecture as the DM. We have the same design as in the discrete control case of the distance metric $\mathit{d}_{\mathcal{M}}$ on space $\mathcal{M}$.  

\subsection{Training Details}
\label{subsec:training}
\textbf{Training Algorithm.}
We describe the main training procedure in Algorithm \ref{algorithm}. Note that for the convenience of description, we parameterize the encoder $f$, dynamics model $h$, backward dynamics model $b$, and policy $\pi$ with $\theta_f$, $\xi_h$, $\xi_b$, and $\omega$, respectively. 
\renewcommand{\algorithmicrequire}{\textbf{Require:}}  
\begin{algorithm}[t]
  \caption{Training Algorithm for \ourname}
  \label{algorithm}
  \begin{algorithmic}[1]
    \Require
      denote parameters of an encoder $f$, a dynamics model $h$, a backward dynamics model $b$ and a policy learning head $\pi$, as $\theta_f$, $\xi_h$, $\xi_b$ and $\omega$, respectively; \\ 
      denote the number of prediction steps as $K$, the number of virtual trajectories as $M$; \\
      denote the prediction loss weight and the predefined maximum weight for cycle consistency loss as $\lambda_{pred}$ and $\lambda_{cyc}^{max}$, respectively;  \\
      denote the warmup end iteration as $i_{end}$;. \\ 
      denote the replay buffer as $\mathcal{D}$;  \\
      denote the interaction step index for Atari and the environment step index  for DMControl as $i$;
     \State randomly initialize all network parameters and make the reply buffer empty.
      \While {$train$}
        \State determine the action $\mathbf{a} \sim \pi(f(\mathbf{s}))$ (based on policy) and interact with environment
        \State record/collect experience  $\mathcal{D}\leftarrow \mathcal{D}\cup (\mathbf{s},\mathbf{a},\mathbf{s}_{next},r)$
        \State sample a sequence of $(\mathbf{s},\mathbf{a},\mathbf{s}_{next},r) \sim \mathcal{D}$
        \State $\mathcal{L}_{cyc} \leftarrow 0$; $\mathcal{L}_{pred} \leftarrow 0$; $\mathcal{L}_{rl} \leftarrow 0$
        \State $\mathbf{z}_t \leftarrow f(\mathbf{s}_t)$
        \For {$j=1,2,...,M$}
            \State $\{\check{\mathbf{a}}_{t}^{(j)}, \check{\mathbf{a}}_{t+1}^{(j)}, \dots, \check{\mathbf{a}}_{t+K-1}^{(j)}\} \sim \mathcal{A}$ \algorithmiccomment{randomly sample a sequence of actions}
            \State $\hat{\mathbf{z}}_t^{(j)} \leftarrow \mathbf{z}_t$
            \For {$k=0,1,...,K-1$}
                \State $\hat{\mathbf{z}}^{(j)}_{t+k+1} \leftarrow h(\hat{\mathbf{z}}^{(j)}_{t+k},\check{\mathbf{a}}^{(j)}_{t+k})$ 
                \algorithmiccomment{(forward) dynamics prediction}
            \EndFor
            
            \State $\mathbf{z}'^{(j)}_{t+K} \leftarrow \hat{\mathbf{z}}^{(j)}_{t+K}$
            \For {$k=K-1,K-2,...,0$}
                \State $\mathbf{z}'^{(j)}_{t+k} \leftarrow b(\mathbf{z}'^{(j)}_{t+k+1},\check{\mathbf{a}}^{(j)}_{t+k})$ \algorithmiccomment{backward dynamics prediction}
            \EndFor
            \State $\mathcal{L}_{cyc} \leftarrow \mathcal{L}_{cyc}+d(\mathbf{z}'^{(j)}_{t}, \mathbf{z}^{(j)}_{t})$ \algorithmiccomment{calculate cycle-consistency loss}
        \EndFor
        \State $\mathcal{L}_{cyc} \leftarrow \mathcal{L}_{cyc} / M$
        \State calculate the forward prediction loss $\mathcal{L}_{pred}$ according to Eq.~(2)
        \State calculate the RL loss $\mathcal{L}_{rl}$
        \State warmup $\lambda_{cyc}$ based on $\lambda_{cyc}^{max},i_{end}, i$
        \State $\mathcal{L}_{total} \leftarrow \mathcal{L}_{rl}+{\lambda_{pred}}\mathcal{L}_{pred}+{\lambda_{cyc}}\mathcal{L}_{cyc}$
        \State $\theta_f, \xi_h, \xi_b, \omega \leftarrow Optimize((\theta_f, \xi_h, \xi_b, \omega),\mathcal{L}_{total})$
    \EndWhile
  \end{algorithmic}
\end{algorithm}

\textbf{Hyperparameters.} We present the hyperparameters used for benchmarks of Atari and DMControl in Table \ref{table:atari_hyperparam} and \ref{table:dmc_hyperparam}, respectively.
We set them mainly following SPR~\cite{schwarzer2021dataefficient} on Atari, and CURL~\cite{laskin2020curl} on DMControl. 

\textbf{Loss Details.} Our total loss \tcb{is composed of} three components: RL loss $\mathcal{L}_{rl}$, prediction loss $\mathcal{L}_{pred}$ and cycle loss $\mathcal{L}_{cyc}$. The RL loss is only applied on real trajectories to update the encoder and the policy learning head. The prediction loss is applied on real trajectories \tcr{to update} the encoder and the DM. \tcr{The cycle consistency loss acts only} on virtual trajectories \tcr{to update} the encoder, the DM and the BDM. Note that we experimentally \tcb{observe that additionally applying the cycle consistency loss on the real trajectories achieves \tcr{only} slight further improvement. For example, it achieves 0.1\% improvement on Atari in the median human-normalized score (\ieno, median HNS).}

\tcr{\textbf{Warmup Scheme.}} In the early stage of training, the dynamics model has not been trained well and thus the cycle-consistency constraint may not be reliable. 
Therefore, inspired by \cite{DBLP:conf/iclr/LaineA17,qiao2018deep}, we ramp up the weight $\lambda_{cyc}$ for the cycle-consistency loss from a small number close to $0$ to a maximum number $\lambda_{cyc}^{max}$. $i$ denotes the index of interaction step for Atari and the index of environment  step for DMControl. When $i$ is smaller than $i_{end}$, $\lambda_{cyc} = \lambda_{cyc}^{max} \cdot \exp(-5 \cdot (1-\frac{i}{i_{end}})^2)$ according to a Gaussian ramp-up curve before a warmup end iteration $i_{end}$. Otherwise, $\lambda_{cyc} = \lambda_{cyc}^{max}$. We set $i_{end}$ to 50k. 
We set $\lambda_{pred}=1$ and $\lambda_{cyc}^{max}=1$. 

\tcr{\textbf{GPU Setup.}} In this work, we run each experiment on one GPU (NVIDIA Tesla V100, P40 or P100).

\subsection{Environment and Code}
In this work, we evaluate models on Atari \cite{bellemare2013arcade} and DMControl \cite{tassa2018deepmind}, which are commonly used benchmarks for discrete and continuous control, respectively. The two benchmarks do not involve personally identifiable information or offensive contents. Our implementation code for Atari is based on SPR \cite{schwarzer2021dataefficient} assert\footnote{Link: \url{https://github.com/mila-iqia/spr}, licensed under the MIT License.}, and that for DMControl is mainly based on CURL \cite{laskin2020curl} assert\footnote{Link: \url{https://github.com/MishaLaskin/curl}, licensed under the MIT License.}.

\subsection{Error Bar of Main Results}
Due to space limitation, we report the error bar (the mean and standard deviation over 10 random seeds) only on DMControl-100k and report the mean scores on Atari-100k. Here, we report the standard deviation over 15 random seeds for both \emph{Baseline}~(\ieno, SPR run by us) and \ourname~on Atari-100k in Table \ref{table:atari_std}. We can see that the standard deviation of our \ourname~is comparable with that of \emph{Baseline}.
\begin{table*}[h]
    \centering
    \caption{The standard deviation (STD) comparison of \textit{Baseline} and \ourname~on Atari-100k. The STD is obtained from 15 runs with random seeds.}
    \label{table:atari_std}
    \scalebox{0.7}{
        \begin{tabular}{l c c l c c l c c}
            \toprule
            \textbf{Game} & \textbf{Baseline} & \textbf{\ourname} & \textbf{Game} & \textbf{Baseline} & \textbf{\ourname} & \textbf{Game} & \textbf{Baseline} & \textbf{\ourname} \\
            \midrule
            Alien & 138.8 & 231.7 & Crazy Climber & 6275.9 & 4664.4              & Kung Fu Master & 4095.1 & 6198.7   \\
            Amidar & 43.0 & 41.3 & Demon Attack & 207.6 & 332.4                 & Ms Pacman & 546.9 & 330.7 \\
            Assault & 138.8 & 50.2 & Freeway & 15.3 & 13.9                      & Pong & 6.5 & 13.2  \\
            Asterix & 229.8 & 170.5 & Frostbite & 1075.0 & 1196.3               & Private Eye & 0.0 & 23.5  \\
            Bank Heist & 97.2 & 160.9 & Gopher & 251.9 & 276.6                  & Qbert & 1053.2 & 952.6 \\
            Battle Zone & 4027.3 & 5261.6 &  Hero & 2940.3 & 2130.9             & Road Runner & 3940.8 & 3765.5\\
            Boxing & 13.6 & 19.9 &  Jamesbond & 47.3 & 75.3                     & Seaquest & 111.9 & 126.9\\
            Breakout & 3.9 & 4.4 & Kangaroo & 3551.8 & 3183.0                   & Up N Down & 2848.4 & 10398.1\\
            Chopper Command & 337.0 & 318.7 & Krull & 323.7 & 524.6 &&& \\
            \bottomrule
        \end{tabular}
    }
\end{table*}

\section{More Experimental Results and Analysis} \label{more_experiments}

\subsection{More Ablation Studies} \label{subsec:more_ablations}
We present more ablation studies, including effectiveness of \ourname~at different environment steps, warmup scheme, weight for cycle consistency loss and where to add the cycle consistency constraint. We use the median HNS of the 26 Atari games and the median score of the 6 DMControl environments to measure the overall performance on Atari and DMControl, respectively. We run each game in Atari with 15 random seeds. To save computational resource, we run each environment in DMControl with 5 random seeds.

\begin{figure*}[t]
	\begin{center}
		\includegraphics[scale=0.5]{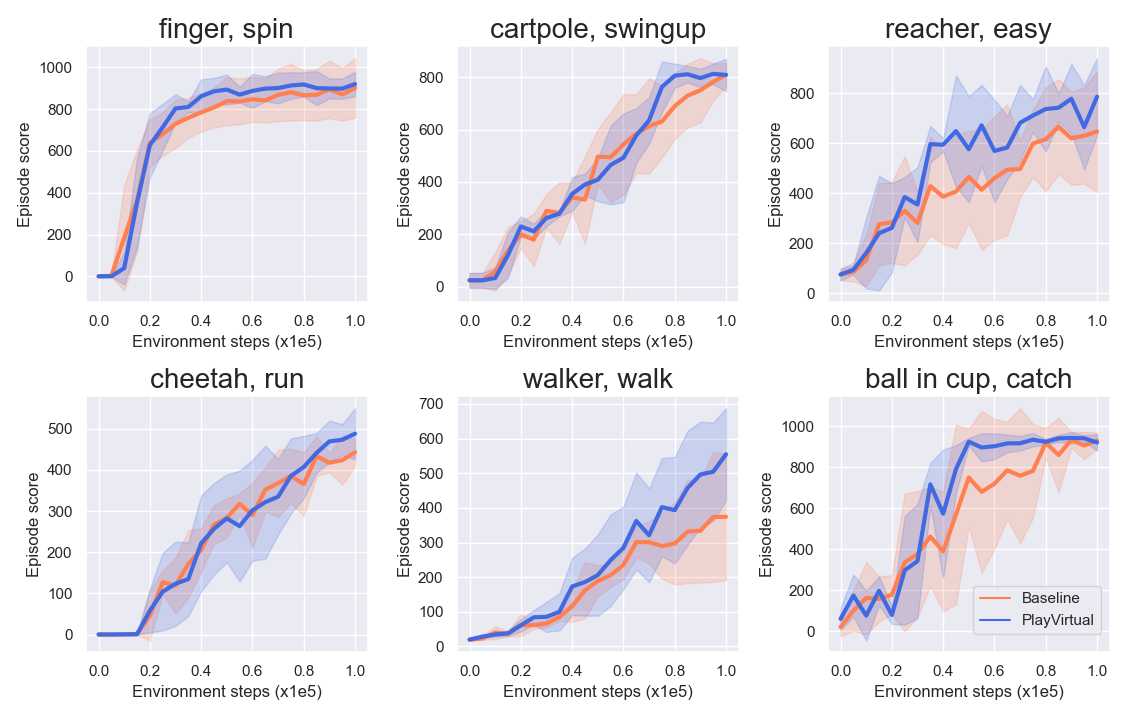}   
	\end{center}
	\caption{Test performance comparison on DMControl where the lines denote the mean score  and the shadow indicates the corresponding standard deviation (obtained by running each environment with 5 random seeds). Our \ourname~(marked with blue) outperforms \emph{Baseline}~(marked with orange) in most environments by a large margin at different environment steps. 
	}
	\label{fig:test100k}
\end{figure*}
\textbf{Effectiveness of \ourname~at Different Environment Steps.}
To further benchmark \ourname's data efficiency, we compare the testing performance in every 5k environment steps at the first 100k on DMControl. Figure \ref{fig:test100k} shows the test performance curves of \textit{Baseline} (SPR\bmark) and \ourname. We can see that our \ourname~performs better than \textit{Baseline} in most environments, where the curves of \ourname~outperform \textit{Baseline} by a large margin on "reacher, easy", "walker, wall", and "ball in cup, catch" environments.

\textbf{Effectiveness of the Warmup for $\lambda_{cyc}$.} Instead of setting $\lambda_{cyc}$ to be a predefined value $\lambda_{cyc}^{max}$, as described in Appendix \ref{subsec:training}, we ramp up the weight $\lambda_{cyc}$ in training.
We compare the performance of our \ourname~without using warmup and with warmup in Table \ref{table:ablation_warm}, which shows that warmup can benefit the training and results in better performance.
\begin{table*}[h]
    \centering
    \caption{Influence of warmup for the weight $\lambda_{cyc}$ w.r.t. the cycle consistency loss.}
    \label{table:ablation_warm}
    \scalebox{1}{
        \begin{tabular}{l c c c }
            \toprule
            \textbf{Model} & $\textbf{Atari-100k}$ & $\textbf{DMControl-100k}$\\
            \midrule
            Baseline & 37.1 & 728.0 \\
            \ourname(w/o warmup) & 42.5 & 749.5 \\
            \ourname & \textbf{47.2} & \textbf{797.0} \\
            \bottomrule
        \end{tabular}
    }
\end{table*}

\textbf{Influence of Predefined Weight $\lambda^{max}_{cyc}$ w.r.t. the Cycle Consistency Loss.}
We set a maximum weight value $\lambda^{max}_{cyc}$ for the cycle consistency loss in the warmup scheme. We study the influence of this hyperparameter in Table \ref{table:ablation_weight}. We find that  $\lambda^{max}_{cyc}=1$ provides superior performance for both Atari and DMControl.
\begin{table*}[h]
    \centering
    \caption{Influence of predefined weight $\lambda^{max}_{cyc}$ w.r.t. the cycle consistency loss.}
    \label{table:ablation_weight}
    \scalebox{1}{
        \begin{tabular}{l c c c c c c}
            \toprule
            $\lambda_{cyc}^{max}$ & 0 & 0.1 & 1 & 2 & 10\\
            \midrule
            Atari-100k & 37.1 & 40.7 & \textbf{47.2} & 45.5 & 41.9 \\
            DMC-100k  & 723.0 & 777.0 & \textbf{797.0} & 740.5 & 763.5 \\
            \bottomrule
        \end{tabular}
    }
\end{table*}

\textbf{Where to Add the Cycle Consistency Constraint?}
For the cycle consistency constraint, we can add this constraint at the end step (\ieno, $d_{\mathcal{M}}(\mathbf{z}'_t, \tilde{\mathbf{z}}_t)$ at $t$) or at every step (\egno, $d_{\mathcal{M}}(\mathbf{z}'_t, \tilde{\mathbf{z}}_t) + \sum_{k=1}^{k=K-1} d_{\mathcal{M}}(\mathbf{z}'_{t+k}, \hat{\mathbf{z}}_{t+k})$) w.r.t.~the backward trajectory (see Figure~1 in our main manuscript for better understanding). Table \ref{table:ablation_location} shows the performance for the two cases. We find their results are similar, where the end-step case is slightly better than the every-step case. 
A possible explanation is that the estimated states from the DM may be not accurate and the supervision from them in every step (besides the end-step) may bring side-effect. For simplicity, we add the cycle consistency constraint only at the end-step where the state $\tilde{\mathbf{z}}_t$ (which is obtained from the \tcb{observation $\mathbf{s}_t$}) is reliable.
\begin{table*}[h]
    \centering
    \caption{Ablation study on where to add the cycle consistency constraint.}
    \label{table:ablation_location}
    \scalebox{1}{
        \begin{tabular}{l c c c }
            \toprule
            \textbf{Model} & $\textbf{Atari-100k}$ & $\textbf{DMControl-100k}$\\
            \midrule
            Baseline & 37.1 & 728.0 \\
            \ourname(every step) & 46.1 & 781.0 \\
            \ourname(end step) & \textbf{47.2} & \textbf{797.0} \\
            \bottomrule
        \end{tabular}
    }
\end{table*}

\subsection{Complexity}
We compare the complexity of \ourname~with \textit{Baseline} in terms of running time and the number of parameters.
The inference time of \ourname~is exactly the same as \textit{Baseline}, since the network architecture of their encoder and the policy learning head are the same, where the auxiliary task is discarded in test. 
Averagely, our method increases \textit{Baseline}'s training time by about 6$\%$ on Atari and 12$\%$ on DMControl, which is acceptable. 

\ourname~introduces a backward dynamics model on top of \textit{Baseline} in training. \ourname~has a very close number of parameters to that of \textit{Baseline} on DMControl.
For example, on "cartpole, swingup" (DMControl), \ourname~has 25.86M parameters while \textit{Baseline} has 25.81M  parameters.
On "pong" (Atari), \ourname~has 3.91M parameters while \textit{Baseline} has 3.83M parameters.

\section{More Discussion} \label{discussion}

\tcr{
\textbf{How Does PlayVirtual Avoid Trivial Solutions in the Latent Space?} Our proposed method \tcb{does} not fall into trivial solutions (such as a constant representation vector) due to the following reasons. (i) We adopt the policy learning (RL) loss to update the encoder to prevent it from falling into this trivial solution. (ii) We also do inference for the dynamics model using real trajectories and supervise the prediction with the \tcb{representations of the} groundtruth states. (iii) We also adopt a target encoder and stop gradient scheme as in SPR \cite{schwarzer2021dataefficient} and BYOL \cite{grill2020byol} to avoid the representation collapse.
}

\tcr{
\textbf{Performance of Dynamics Model.} We conduct an evaluation on \tcb{the} dynamics model (DM)\tcb{. Particularly,} after 100k environment steps training, we calculate the average prediction mean squared error (MSE) of DM in latent space over 1000 transitions. The evaluation is on a subset of DMControl environments with 5 random seeds. The comparison results of \textit{Baseline} (SPR\bmark) and \ourname~are shown\mbox{ in Table \ref{table:discuss_dm}}. We can see that our \tcb{models achieve} better prediction performance than \textit{Baseline}. \tcb{Thanks to our} cycle-consistency regularized virtual trajectories generation, we safely augment the trajectories for learning better state representations\tcb{, which also results in} a stronger dynamics model.
\begin{table*}[h]
    \centering
    \caption{Evaluation on dynamics models in \textit{Baseline} and our method. The mean squared error (MSE) results of dynamics prediction are reported.}
    \label{table:discuss_dm}
    \scalebox{1}{
        \begin{tabular}{l c c c }
            \toprule
            \textbf{MSE} & $\textbf{Cartpole, swingup}$ & $\textbf{Reacher, easy}$ & $\textbf{Cheetah, run}$\\
            \midrule
            Baseline & 0.2517 & 0.3920 & 0.0731 \\
            \ourname & \textbf{0.2357} & \textbf{0.3633} & \textbf{0.0672} \\
            \bottomrule
        \end{tabular}
    }
\end{table*}
}

\tcr{
\textbf{Performance of Learned Representations.} Besides the final performance reported in our main manuscript, we further evaluate the state representations by studying which kind of representations can better promote the policy learning. 
\tcb{As shown in Table \ref{table:discuss_rep}, we consider three schemes. (i) For \textit{None}, models are trained from scratch with only RL loss (\ieno, $\mathcal{L}_{rl}$). (ii) For \textit{Baseline Encoder}, models are trained with only RL loss while their encoders are initialized with (100k environment steps) SPR\bmark-pretrained encoder parameters, and these encoders are fixed during training. (iii) For \textit{\ourname~Encoder}, the setting is similar to (ii) except for initializing the encoders with \ourname-pretrained encoder parameters. We test the 100k-step performance (\ieno, scores) on a subset of DMControl environments with 5 random seeds. As shown in Table \ref{table:discuss_rep}, we can observe that the model whose encoder is initialized by a pretrained \textit{\ourname~Encoder} performs better than that of \textit{Baseline Encoder} and non-pretrained non-fixed encoder (\ieno, \textit{None}). This observation demonstrates the state representations learned by our method are more helpful to the policy learning.}
\begin{table*}[h]
    \centering
    \caption{Evaluation on learned representations. The 100k-step scores of models with different pretrained encoders are reported.
    }
    \label{table:discuss_rep}
    \scalebox{1}{
        \begin{tabular}{l c c c }
            \toprule
            \textbf{Initialization} & $\textbf{Cartpole, swingup}$ & $\textbf{Reacher, easy}$ & $\textbf{Cheetah, run}$\\
            \midrule
            None & 796 $\pm$ 60 & 730 $\pm$ 185 & 388 $\pm$ 89 \\
            Baseline Encoder & 839 $\pm$ 24 & 517 $\pm$ 141 & 478 $\pm$ 30 \\
            \ourname~Encoder & \textbf{847 $\pm$ 31} & \textbf{828 $\pm$ 67} & \textbf{512 $\pm$ 31} \\
            \bottomrule
        \end{tabular}
    }
\end{table*}
}

\tcr{
\textbf{Method of Action Sampling.} In this work, we \tcb{uniformly} sample actions from the action space when generating virtual trajectories. Although the study of action sampling is not the focus of this work, we do evaluate other action sampling methods such as adding zero-mean Gaussian noise $\mathcal{N}(0, \sigma)$ to the original actions in the real trajectories. We conduct the experiment with 5 random seeds. The results in Table \ref{table:discuss_act} show that using uniformly sampled actions  (\ieno, \textit{Random Action}) achieves higher performance than the above-mentioned Gaussian-noise perturbed actions (\ieno, \textit{Perturbed Action ($\sigma$)}). This maybe because random actions can "explore" more states for boosting representation learning. Further, there can be more advanced sampling methods such as surprise-based sampling or policy-guided sampling. \tcb{We} leave the study on them as future work.
\begin{table*}[h]
    \centering
    \caption{Study on action sampling methods in generating virtual trajectories. \textit{Perturbed Action ($\sigma$)} denotes adding $\mathcal{N}(0, \sigma)$ Gaussian noise to the original actions, while \textit{Random Action} indicates uniformly sampled actions. We report the median scores across 6 DMControl environments.
    }
    \label{table:discuss_act}
    \scalebox{0.766}{
        \begin{tabular}{l c c c c }
            \toprule
            \textbf{DMControl} & $\textbf{Perturbed Action (0.01)}$ & $\textbf{Perturbed Action (0.02)}$ & \textbf{Perturbed Action (0.05)} &$\textbf{Random Action (Ours)}$\\
            \midrule
            Median Score & 732.0 & 747.0 & 764.0 & \textbf{797.0} \\
            \bottomrule
        \end{tabular}
    }
\end{table*}
}

\tcr{
\textbf{Why Do We Predict Dynamics in the Latent Space?} We predict environment dynamics in the latent space instead of the observation space for two reasons. (i) For high-dimensional control tasks such as image-based RL, we \tcb{expect to} learn compact and informative representations that exclude control-irrelevant information to better serve policy learning. If we stay in the observation space, the representations \tcb{would} include control-irrelevant information to reconstruct some control-irrelevant details, which \tcb{distracts} RL algorithms and slows down the policy learning speed \cite{zhang2021learning}. (ii) Staying in the latent space requires less computational cost as the dimension is \tcb{lower}.
}

\textbf{Application and Limitation.}
Our proposed method \ourname, which augments cycle-consistent virtual trajectories, is generic and can be applied to many existing RL frameworks. In this work, we apply it on top of two model-free methods: SPR for discrete control benchmark and on top of a variant of SPR, \ieno, SPR\bmark for continuous control benchmark. But it is not limited to the two baselines. \tcr{\tcb{Our method should be applicable} to model-based RL methods \tcb{to improve data efficiency}. We leave the implementation on top of other model-free or model-based baselines as future work. However, our method also bears some limitations such as not excelling in non-deterministic environments \tcb{where} the environment dynamics is difficult to be modeled and the cycle consistency in the forward-backward trajectory may be hard to meet.}

\section{Potential Societal Impact} \label{impact}
Deep reinforcement learning (RL) has broad applications, including games, robotics, healthcare, dialog systems, \textit{etc}. Learning good feature representations is important for deep RL. However, with limited experience, RL often suffers from data inefficiency for training. In this work, we propose a general method, dubbed \ourname, which augments cycle-consistent virtual trajectories to enhance the data efficiency for RL feature representation learning. We have demonstrated the effectiveness of our \ourname, which achieves the best performance on both discrete control benchmark and continuous control benchmark. We believe our technique will promote the progress of RL applications and inspire more interesting works on improving the data efficiency for RL.
\tcb{Meanwhile, for image-based RL, systems should be developed following responsible AI policies to be fair and safe.}

\newpage
\begin{table*}[h]
    \centering
    \caption{Hyperparameters used for Atari.}
    \label{table:atari_hyperparam}
    \scalebox{1.0}{
        \begin{tabular}{l c c c c l}
            \toprule
            \textbf{Hyperparameter} & & & & & \textbf{Value} \\
            \midrule
            Gray-scaling & & & & & True \\
            Frame stack & & & & & 4 \\
            Observation downsampling & & & & & (84, 84) \\
            Augmentation  & & & & & Random shift $\&$ intensity \\
            Action repeat & & & & & 4 \\
            
            Training steps & & & & & 100K \\
            Max frames per episode & & & & & 108K \\
            Reply buffer size & & & & & 100K \\
            Minimum replay size for sampling & & & & & 2000 \\
            Mini-batch size & & & & & 32 \\

            Optimizer & & & & & Adam \\
            Optimizer: learning rate & & & & & 0.0001 \\
            Optimizer: $\beta_1$ & & & & & 0.9 \\
            Optimizer: $\beta_2$ & & & & & 0.999 \\
            Optimizer: $\epsilon$ & & & & & 0.00015 \\
            Max gradient norm & & & & & 10 \\

            Update & & & & & Distributional Q \\
            Dueling & & & & & True \\
            Support of Q-distribution & & & & & 51 bins \\
            Discount factor & & & & & 0.99 \\
            Reward clipping Frame stack & & & & & [-1, 1] \\
            
            Priority exponent & & & & & 0.5 \\
            Priority correction & & & & & 0.4 $\rightarrow$ 1 \\
            Exploration & & & & & Noisy nets \\
            Noisy nets parameter & & & & & 0.5 \\
            
            Evaluation trajectories & & & & & 100 \\
            
            Replay period every & & & & & 1 step \\
            Updates per step & & & & & 2 \\
            Multi-step return length & & & & & 10 \\
            
            Q network: channels & & & & & 32, 64, 64 \\
            Q network: filter size & & & & & 8 $\times$ 8, 4 $\times$ 4, 3 $\times$ 3 \\
            Q network: stride & & & & & 4, 2, 1 \\
            Q network: hidden units & & & & & 256 \\
            
            Target network update period & & & & & 1 \\
            $\tau$ (EMA coefficient) & & & & & 0 \\
            \midrule
            \textbf{Additional Hyperparameters in \ourname}& & & & & \\
            \midrule
            K (number of prediction steps)& & & & & 9 \\
            M (number of virtual trajectories)& & & & & $2|\mathcal{A}|$ (two times of action space size) \\
            $\lambda_{pred}$ (weight for prediction loss)& & & & & 1 \\
            $\lambda_{cyc}^{max}$ (a weight related to cycle consistency loss)& & & & & 1 \\
            Warmup & & & & & Gaussian ramp-up ($i_{end}$=50K) \\
            \bottomrule
        \end{tabular}
    }
\end{table*}

\newpage
\begin{table*}[h]
    \centering
    \caption{Hyperparameters used for DMControl.}
    \label{table:dmc_hyperparam}
    \scalebox{1.0}{
        \begin{tabular}{l c c c c l}
            \toprule
            \textbf{Hyperparameter} & & & & & \textbf{Value} \\
            \midrule
            Frame stack & & & & & 3 \\
            Observation rendering & & & & & (100, 100) \\
            Observation downsampling & & & & & (84, 84) \\
            Augmentation  & & & & & Random crop $\&$ intensity \\
            Replay buffer size & & & & & 100000 \\
            Initial exploration steps & & & & & 1000 \\
            Action repeat & & & & & 2 finger-spin and walker-walk;\\
             & & & & & 8 cartpole-swingup; \\
             & & & & & 4 otherwise \\
            
            Evaluation episodes & & & & & 10 \\
            
            Optimizer & & & & & Adam \\
            $(\beta_1, \beta_2) \rightarrow (\theta_f, \xi_h, \xi_b, \omega)$ & & & & & (0.9, 0.999) \\
            $(\beta_1, \beta_2) \rightarrow (\alpha)$ (temperature in SAC) & & & & & (0.5, 0.999) \\
            Learning rate $(\theta_f, \omega)$ & & & & & 0.0002 cheetah-run \\
             & & & & & 0.001 otherwise \\
            Learning rate $(\theta_f, \xi_h, \xi_b)$ & & & & & 0.0001 cheetah-run \\
             & & & & & 0.0005 otherwise \\
            Learning rate $(\alpha)$ & & & & & 0.0001 \\
            Policy batch size $(\theta_f, \omega)$ & & & & & 512 \\
            Auxiliary batch size $(\theta_f, \xi_h, \xi_b)$ & & & & & 128 \\
            
            Q-function EMA $\tau$ & & & & & 0.01 \\
            Critic target update freq & & & & & 2 \\

            Discount factor & & & & & 0.99 \\
            Initial temperature & & & & & 0.1 \\

            Target network update period & & & & & 1 \\
            Target network EMA $\tau$ & & & & & 0.05 \\
            \midrule
            \textbf{Additional Hyperparameters in \ourname}& & & & & \\
            \midrule
            K (number of prediction steps)& & & & & 6 \\
            M (number of virtual trajectories)& & & & & 10 \\
            $\lambda_{pred}$ (weight for prediction loss )& & & & & 1 \\
            $\lambda_{cyc}^{max}$ (a weight related to cycle consistency loss)& & & & & 1 \\
            Warmup & & & & & Gaussian ramp-up ($i_{end}$=50K) \\
            \bottomrule
        \end{tabular}
    }
\end{table*}

\end{document}